\PassOptionsToPackage{numbers}{natbib}
\documentclass{article}
\pdfoutput=1

    \usepackage[final, nonatbib]{neurips_edu_2020}

\usepackage[utf8]{inputenc} 
\usepackage[T1]{fontenc}    
\usepackage{hyperref}       
\usepackage{url}            
\usepackage{booktabs}       
\usepackage{amsfonts}       
\usepackage{nicefrac}       
\usepackage{microtype}      
\usepackage[numbers]{natbib}
\usepackage{graphicx}

\title{Representation Based Complexity Measures for Predicting Generalization in Deep Learning}

\author{%
  Parth Natekar \\
  Department of Engineering Design\\
  Indian Institute of Technology, Madras\\
  \texttt{parthnatekar26@gmail.com} \\
   \And
   Manik Sharma \\
  Department of Engineering Design\\
  Indian Institute of Technology, Madras\\
  \texttt{shmakn99@gmail.com} \\
}

\begin{document}

\maketitle

\begin{abstract}
Deep Neural Networks can generalize despite being significantly overparametrized. Recent research has tried to examine this phenomenon from various viewpoints and to provide bounds on the generalization error or measures predictive of the generalization gap based on these viewpoints, such as norm-based, PAC-bayes based, and margin-based analysis.
In this work, we provide an interpretation of generalization from the perspective of quality of internal representations of deep neural networks, based on neuroscientific theories of how the human visual system creates invariant and untangled object representations. Instead of providing theoretical bounds, we demonstrate practical complexity measures which can be computed ad-hoc to uncover generalization behaviour in deep models. We also provide a detailed description of our solution that won the NeurIPS competition on Predicting Generalization in Deep Learning held at NeurIPS 2020. An implementation of our solution is available at \url{https://github.com/parthnatekar/pgdl}.
\end{abstract}

\section{Introduction}

The generalization ability of deep neural networks despite having more parameters than training data is a property of great interest from both theoretical and application perpectives. Understanding the factors behind generalization and possibly creating practical measures which are predictive of generalization can lead to better designs as well as a better understanding of this phenomanon.

Multiple hypothesis have been presented before based on various properties of deep neural networks, from optimization based theories, to empirical measures based on sharpness, to theoretically motivated generalization bounds on the test error based on some intrinsic property of deep networks. 

Theoretically motivated techniques which aim to relate generalization bounds to measures of capacity such as VC Dimension and Rademacher Complexity are vacuous when the network falls under the ‘over-parameterized’ regime. Bartlett \citep{bartlett1998sample} provide a bound based on the norm of the weights, and Neyshabur et al. (2015) \citep{neyshabur2015norm} extend this to other norm based measures. Multiple studies have related generalization to PAC-Bayes based bounds \citep{dziugaite2017computing, langford2001not}. Optimization based theories rely on the presence of implicit regularization via SGD or explicit regularization via dropout or weight decay \citep{neyshabur2014search, srivastava2014dropout}. However, Zhang et al. \citep{zhang2016understanding} showed that despite such implicit and explicit regularization, deep neural networks can achieve negligible training error on randomly labelled samples.

Empirical measures have also been proposed before. Keskar et al. \citep{keskar2016large} connect generalization to flatness of minima, Arora et al. \citep{arora2018stronger} derive a bound based on the noise attenuation properties of deep networks, while Morcos et al. \citep{morcos2018importance} hypothesize that networks' reliance on single directions is a good predictor of generalization performance, and demonstrate this via ablation studies and addition of noise. Jiang et al. (2018) \citep{jiang2018predicting} demonstrate that the margin distribution based measures show a high correlation with generalization gap. 

Many of the theoretical measures discussed above are upper bounds and are not tight enough to use in an empirical setting. Measures based on the weight norms cannot be used when there are variations in network topology. Other measures like PAC-Bayes or empirical measures like sharpness are stochastic and it is not clear how these scale with changes in network size and other network parameters.  Thus, most of these measures do not always provide intuitive support to practitioners in terms of designing models which generalize better or comparing generalization ability of already trained networks with limited access to their hyperparameters and design choices. 

In this work, we provide an interpretation of generalization ability in terms of the quality of representations deep neural networks. We propose simple and intuitive complexity measures that can be computed in a post-hoc setting, i.e. when comparing models after they have already been trained. Our measures are predictive of generalization irrespective of changes in hyperparameters and model architectures, and achieve high scores on the final and development tasks of the NeurIPS competition on Predicting Generalization in Deep Learning. 


\section{Quality of Representations and Neuroscientific Parallels}

Representation learning ascribes the effectiveness of deep learning to the ability of deep networks to learn representations that are invariant to nuisances in the input such as translations, rotations, occlusions, and also “disentangled”, that is separating out independent factors in the high-dimensional space of data \citep{achille2018emergence}. Bengio et al. \citep{bengio2013representation} discuss, broadly, the properties of good feature representations and discuss how general priors about the world around us can be reflected through these. On a high level, it makes sense that deep models which build 'better' feature representations are more likely to generalize better. We posit that certain properties of learned representations in deep neural networks improve generalization ability. However, not all properties that are hypothesized to be linked with representation quality may be positively correlated with generalization, and if the are, may not be easily computable. Moreover, while it is intuitively understood and often taken for granted that representation quality and generalization are linked, there have been limited studies which explicitly relate the two and empirically investigate this relationship.

Neuroscientific and psychological studies have shown that humans rely on invariant and consistent feature representations irrespective of controlled naturalistic variations of objects \citep{karimi2017invariant, nielsen2008object}. Studies have shown that one of the key aspects of invariant object detection is the ability of the visual hierarchy to create easily separable representation manifolds \cite{dicarlo2007untangling}. Robustness is another related and well-known quality of human vision that helps in invariant object detection. Human vision has been shown to be more robust than deep neural networks on image manipulations like contrast reduction, additive noise or eidolon-distortions \citep{geirhos2017comparing}. 

While it is unclear if deep feed-forward convolutional neural networks use similar strategies, studies have shown a hierarchical correspondence between deep network representations and spatio-temporal human visual brain representations \citep{cichy2016comparison}. Thus, creating measurable metrics that quantify the properties of representations discussed above can potentially predict generalization in deep learning.


Based on such neuroscientific arguments and the above ideas from representation learning, we choose three metrics of quality of representations that have the potential to predict generalization in a post-hoc diagnostic setting. These are: (i) Consistency, (ii) Robustness (to controlled variations), and (iii) Separability.

In the next sections, we detail simple complexity measures based on these metrics of quality of representations that allow us to compare models that have been trained with a range of different hyperparameters and architectures. 
We report the performance of our complexity measures on a range of datasets and models which are part of the NeurIPS competition on Predicting Generalization in Deep Learning (PGDL). These complexity measures achieve top scores on the development and final phases of PGDL. We report our scores on the public and private datasets of PGDL, along with scores of the different techniques we implemented that were not submitted for evaluation.

\section{Preliminaries and Notation}

First, we define some notation. A feedforward neural network is denoted by $f_w: X \rightarrow \mathbb{R}^\kappa$ and its weight parameters by $w$. The weight tensor of the $i^{th}$ layer of the network is denoted by $W_i$, so that $w =$vec$(\mathcal{W}_1, . . . ,\mathcal{W}_d)$, where d is the depth of the network, and vec represents the vectorization operator. Let $\mathcal{A} = (\mathcal{A}_1, . . . , \mathcal{A}_d)$ be the feature maps of the network at each layer, and $f_{w, k}: A_k \rightarrow \mathbb{R}^\kappa$ the intermediate model that maps representations at layer $k$ to the network output. Let $\mathcal{D}$ be the data distribution over inputs and outputs. Let $S = \{X, y\}$ be the training set, where $X = \{X_i\}_{i=1}^N$ are the data points and $y = \{y_i\}_{i=1}^N \in \{1, . . . ,\kappa \}$ are the corresponding training labels sampled randomly from $\mathcal{D}$. We denote by $f_w(X)[j]$ be the $j^{th}$ output of the network. 
The empirical 0-1 classification loss $L$ is defined as $\hat{L} = \frac{1}{m} \sum_{i = 1}^m  I\big(f_w(\tilde{X}_i)[y_i] \leq \max_{j \neq \tilde{y}_i} f_w(\tilde{X}_i)[j] \big)$. 
Finally, the generalization gap is defined as the difference between the training and test accuracy of the network.

Our task is to find a complexity measure $\mathcal{C}$ that is predictive of the generalization gap using only the network weights and the input dataset in a post-hoc setting, i.e. after models have been trained. Such a measure should allow a relative comparison between models trained using a range of different hyperparameters and architectures, thus capturing properties of deep neural networks that are responsible of generalization irrespective of how they are trained.

In order to better capture relationships between complexity measures and the generalization gap and to separate these from spurious correlations, a conditional mutual information based metric is used as in Jiang et al. (2019) \citep{jiang2019fantastic}.

\section{Representation Based Complexity Measures}

\subsection{Consistency of Representations}

It is logical that networks which build consistent representations of similar objects are good at generalizing. In such a case, it is likely that the network is recognizing the underlying causal factors that define that object, which allows it to create consistent representations in-spite of variations. 

\begin{figure}
    \centering
    \includegraphics[width = 1.\textwidth]{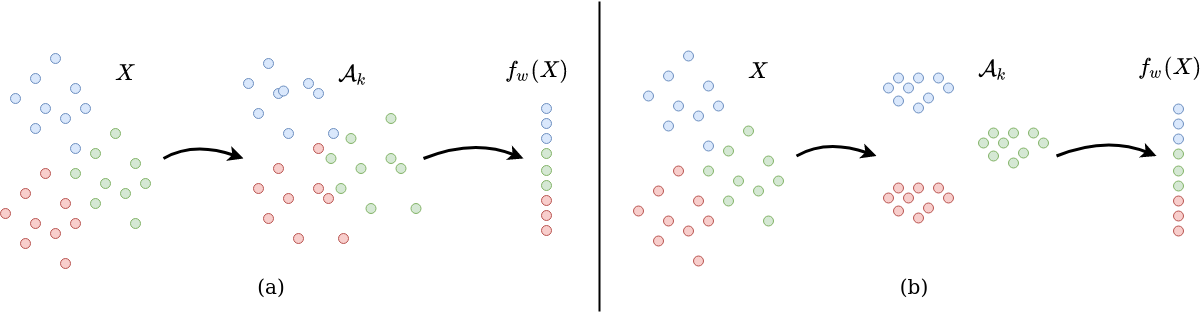}
    \caption{A visual depiction of how consistency of representations could aid in generalization. Models in both Figure \ref{fig:fig 1}(a) and Figure \ref{fig:fig 1}(b) are saturated, and thus correctly classify the training data $X$. However, the model in Figure \ref{fig:fig 1}(b) has more consistent representations. Calculating the DB Index of intermediate representations $A_k$ can then be predictive of generalization gap.}
    \label{fig:fig 1}
\end{figure}

Clustering of feature representations to encourage compactness and interpretability and to create parsimonious representations has been explored before \citep{liao2016learning}. Clustering is also used in the context of network compression \citep{son2018clustering}. In the context of generalization, measuring how clustered representations of a particular class are tell us about their consistency and compactness. Models where feature representations at layer $L_i$ for each class lie in their own cluster and are easily distinguishable from other class clusters in high dimensional feature spaces are likely to have similar representations at subsequent layers $L_{j | j > i}$, and hence likely to be predicted to be in the same class. Figure \ref{fig:fig 1} provides a visual depiction of this idea.

Therefore, we require a measure that tells us about the quality of the clustering or compactness of feature representations in intermediate layers of the model, given the ground-truth label as the cluster index. However, simply measuring the compactness of representations using distance-based measures does not suffice when comparing models with different topology. For example, one possible metric could be the expectation over all classes of the average distance of representations of a class to the cluster centroid of that class. The smaller the average distance to cluster centroids for each class, the more consistent the representations. However, this is likely to fail when models have different shapes, and comparisons happen across different vector spaces. We use the Davies-Bouldin Index to measure clustering quality. The Davies-Bouldin Index measures the ratio of within-cluster scatter to between cluster separation. Being a ratio, it is easily comparable across models with different architectures. Measuring the Davies-Bouldin index on intermediate feature representations tells us how consistent representations are within a class and how different they are from other classes. Mathematically, for a given layer $L_k$ and its activations $A_k$,

\begin{equation}
    \tilde{\mathcal{A}}_k = \Phi(\mathcal{A}_k)
\end{equation}
\begin{equation}
    \mathcal{S}_i = \Big( \frac{1}{n_i} \sum_{1}^{n_i} | \tilde{\mathcal{A}}_k^i - \mu_{\tilde{\mathcal{A}}_k^i} |^p \Big)^{1/p}
\end{equation}
\begin{equation}
    \mathcal{M}_{i, j} = || \mu_{\tilde{\mathcal{A}}_k^i} - \mu_{\tilde{\mathcal{A}}_k^j} ||_p 
\end{equation}

Where, $\Phi$ indicates a dimensionality reduction operation, $i$ and $j$ are two different classes, $A_k^i$ are the feature representations of samples belonging to class $i$, $\mu_{\tilde{\mathcal{A}}_k^i}$ is the cluster centroid of the representations of class $i$, $S_i$ is a measure of scatter within representations of class $i$, and $\mathcal{M}_{i, j}$ is a measure of separation between representations of classes $i$ and $j$. Then, the complexity measure based on the Davies Bouldin Index for representations at layer $k$ is calculated as:

\begin{equation}
    \mathcal{C}_k = \frac{1}{\kappa} \sum_{i = 1}^{\kappa} \max_{i \neq j} \frac{\mathcal{S}_i + \mathcal{S}_i}{\mathcal{M}_{i, j}}
\end{equation}

In practice, we compute the Davies-Bouldin Index over minibatches of data and average the result to get the final complexity measure. We choose a suitable batch size based on the number of output classes such that there are a reasonable number of samples from each class in a batch. We also find that taking the mean as opposed to the maximum over clusters gives better results on most tasks, possibly because this quantifies the separation of a class from all other classes as opposed to only the closest cluster.

We explore PCA and Max-Pooling for dimensionality reduction. Also, the Davies-Bouldin Index could be computed for any intermediate layer. It makes sense that middle and deeper layers of the network are most suitable for computing the DB Index; these are likely to capture high level concepts in a compact fashion. Initial layers are likely to detect edges and simple shapes and thus have high variance, while the penultimate layer before the output may have converged representations that are consistent irrespective of generalization gap. We find that computing the DB Index on middle and deeper layers before the penultimate layer gives highest scores for the public tasks. However, from the perspective of comparing models with different topologies, we find that the DB Index on the first layer representations is most suitable. We also explore other measures of clustering quality, such as Silhouette Coefficient and Calinski-Harabasz Index but find that the Davies Bouldin Index works best.

\subsection{Robustness of Representations}

Multiple studies have found the human visual system to be more robust to most types of noise than deep neural networks \citep{geirhos2017comparing, geirhos2018generalisation}. However, it has also been shown that networks trained on perturbed images perform better than humans on the exact perturbation types they were trained on, but generalize poorly on other perturbation types \citep{geirhos2018generalisation}. Robustness of internal network representations is thus closely related to generalization ability and a metric which quantifies this could be predictive of generalization.

Supervised Deep Learning approaches are usually trained on the principle of Empirical Risk Minimization, where the average error over the training samples is minimized. Vicinal Risk Minimization, the formal name for data augmentation and related techniques, aims to reduce the error on samples in the vicinity of the training samples. 

Measuring error on a vicinal dataset is one seemingly straightforward way to predict performance of models on test data. This would allow us to determine how robust a model is to nuisance or perturbations in features. However, multiple problems arise here. One, data augmentation or perturbation techniques are often dataset dependent. Creating a vicinal distribution requires visualization and expert understanding of the underlying factors of variation of the data, which is possible for natural image datasets but may not be for the range and types of data deep neural networks are being used for today. Secondly, such a vicinal distribution would only allow us to check robustness in the input space. Thirdly, in a post-hoc diagnostic setting, it is necessary to know whether networks are trained on a vicinal distribution, and if they are, what sort of perturbations have they been trained to handle. Checking performance of a model on a vicinal distribution may not tell us anything about the test performance if our chosen perturbations take the data into a neighbourhood that the model has never seen; all models are likely to give out random results here, irrespective of their test performance. This is especially likely in higher dimensional spaces, where it is easy to be close to the training distribution and yet be in an unseen region. Moreover, comparing models trained on vicinal distributions and those trained on the original training dataset becomes difficult here.

Thus, while predicting generalization in a post-hoc setting, perturbations to the training dataset or the intermediate representations have to be (i) data-agnostic, and (ii) controlled, to ensure that they do not take representations into a space that the model has never seen before.

Mixup \citep{zhang2017mixup} and manifold mixup \cite{verma2019manifold} provide a potential solution that meets our requirements. Mixup is data agnostic, and perturbs representations in directions that our model has already seen, i.e. towards another training sample. 

However, mixup by itself is an ante-hoc procedure which trains models to behave linearly between samples irrespective of class. If models have not been trained on mixup, it is likely that models will perform poorly on mixed-up samples irrespective of how well they generalize. We use a label-wise version of mixup where we check performance on linear combinations of training samples restricted to a class. Models are more likely to have linear behaviour within samples of a particular class as opposed to samples of different classes, especially if they haven't been trained with mixup. We find that our variation of mixup is especially useful in a post-hoc setting and achieves significantly higher scores on all tasks of PGDL.
Mathematically, we create a vicinal distribution $\mathcal{V}$ made up of tuples $\{\tilde{X}, \tilde{y}\}$, such that

\begin{equation}
    \tilde{X}^i = \lambda X_1^i + (1-\lambda) X_2^i
\end{equation}
\begin{equation}
    \tilde{y}^i = y^i
\end{equation}

More generally, we create mixed-up representations $\tilde{A}$ such that,
\begin{equation}
    \tilde{A}^i = \lambda A_1^i + (1-\lambda) A_2^i
\end{equation}
\begin{equation}
    \tilde{y}^i = y^i
\end{equation}
Where inputs $X$ are denoted by $A_0$. Computing model performance on mixed-up representations can then tell us about its generalization ability. The mixup complexity measure based on representations at layer $k$ is then defined as:

\begin{equation}
    \mathcal{C}_k = \sum_{i=1}^\kappa \frac{1}{N_i} \sum_{n = 1}^{N_i}  I\big(f_{w, k}(\tilde{A}_n^i)[y_i] \leq \max_{j \neq \tilde{y}_n} f_{w, k}(\tilde{A}_n^i)[j] \big)
\end{equation}
\begin{equation}
     = \sum_{i=1}^\kappa \hat{L}\big(f_{w, k}(\tilde{A}^i)\big)
\end{equation}

\begin{figure}
    \centering
    \includegraphics[width = 0.5\textwidth]{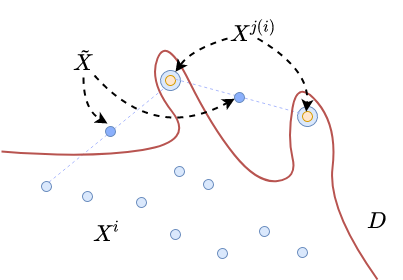}
    \caption{A visualization of how mixup can help quantify random labelling. $X^i$ depicts samples of class $i$, $X^{j(i)}$ depicts samples of class $j$ wrongly labelled as class $i$, $\tilde{X}$ depicts the mixed-up samples, and $D$ shows the hypothetical margin (which overfits on the wrongly labelled samples). The mixed-up points $\tilde{X}$ are likely to be misclassified when either one or both samples in the linear combination are randomly labelled.}
    \label{fig:fig 2}
\end{figure}

Where $f_{w, k}(\tilde{A})$ denotes the model mapping intermediate representations to the output. We report scores for both mixup and manifold mixup on various layers.  However, we find that mixup on the input data is most predictive of generalization. Mixup also has the added advantage of accounting for random labelling. Networks are likely to misclassify linear combinations of input samples where atleast one sample has been randomly labelled. This is shown graphically in Figure \ref{fig:fig 2}. Thus, performance on mixed-up samples can also be used as a proxy for the percentage of random labels the network has been trained on.

\subsection{Separability of Representations}

The entanglement of object manifolds is hypothesized to be one reason why object recognition is difficult for humans. It has been proposed that the human visual hierarchy untangles object manifolds to transform inseparable manifolds into those which can be linearly or easily separated \citep{dicarlo2007untangling}, behaviour which has also been seen in deep neural networks \citep{cohen2020separability}. 

Approximate formulations for the margin, the minimum distance to a decision boundary between two classes, have been proposed before for deep neural networks. Margin distributions provide a reasonable proxy for measuring separability of representation manifolds in high dimensional spaces \citep{elsayed2018large}. The margin distribution has also been shown to be empirically predictive of generalization in deep neural networks \citep{jiang2018predicting}. We use the margin distribution as a measure of separability of representations in intermediate layers of the network. Jiang et al. (2018) \citep{jiang2018predicting} show that the normalized margin distribution becomes heavier tailed and shifts to the right as generalization gap increases, and train a linear classifier to predict generalization gap from a signature of the margin distribution. It may not always be possible to train such a classifier in a real world setting. We replace the classifier simply by a measure of central tendency of the margin distribution as a summary measure and find that this gives good results. Moreover, we argue that the margin distribution computed on a perturbed representation manifold or the vicinal distribution is more useful as a predictor of generalization. Mathematically, Let $D_{(i, j)}$ be the decision boundary between two classes $i$ and $j$, and $X'$ be a perturbed input sample. Let $A'_k$ be the corresponding intermediate layer representation. Then, the margin is defined as: 

\begin{equation}
    D_{(i, j), k} = \{A'_k \; | \; f_{w, k}(A'_k)[i] = f_{w, k}(A'_k)[j]\}
\end{equation}
And the margin distance is approximated as:
\begin{equation}
    d_{f_{w,k}, (i, j)}(A'_k) = \frac{f_{w, k}(A'_k)[i] - f_{w, k}(A'_k)[j]}{|| \nabla_{A'_k} f_{w, k}(A'_k)[i] - \nabla_{A'_k} f_{w, k}(A'_k)[j] ||_2 }
\end{equation}

The margin complexity measure at layer $k$ is then:
\begin{equation}
    \mathcal{C}_k = - \theta(d_{f_{w,k}, (i, j)}(A'_k))
\end{equation}

\begin{figure}
    \centering
    \includegraphics[width = 1.\textwidth]{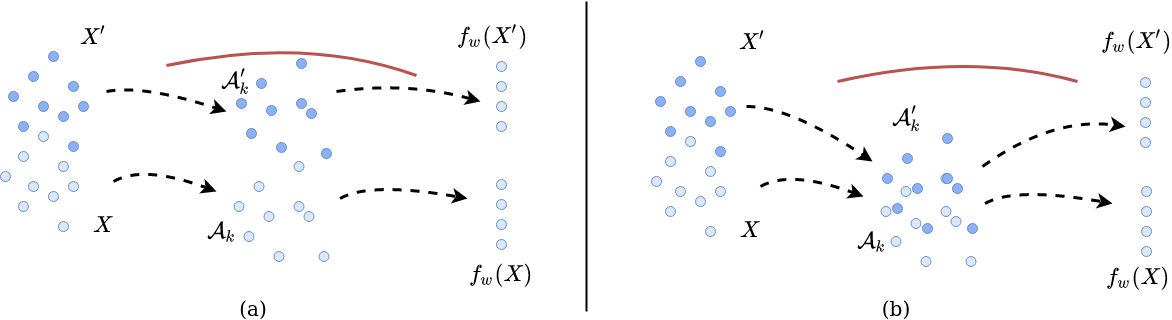}
    \caption{Calculating the margin on the perturbed samples tells us more about generalization ability than the naive margin. The margin is shown in red, $X$ depicts samples from the original distribution, and $X'$ are samples from the vicinal distribution.}
    \label{fig:fig 3}
\end{figure}

Where $\theta$ is a distribution summary measure. We normalize the margin by the total variation of the representation as in Jiang et al. (2018) \citep{jiang2018predicting}. The intuition behind calculating the margin on a perturbed distribution is shown in Figure \ref{fig:fig 3}. Figure \ref{fig:fig 3}(a) and Figure \ref{fig:fig 3}(b) show two models that correctly classify samples $X$ from the original distribution as well as samples $X'$ from a perturbed distribution. The original data points $X$ are at a similar distance from the margin for both models. However, the model in Figure \ref{fig:fig 3}(b) also has a high margin for perturbed samples $X'$. Thus, it is expected that this model will better handle a wider range of data and therefore generalizes better. Note that simply calculating the margin on samples from the original distribution will not help here  (since this will be similar for both models), neither will trivially comparing classification performance or model confidence on perturbed samples (since both models correctly classify all points; we also show this empirically by comparing Augment Margin and simple Augment Performance, shown in Table \ref{tab: table 1}). We explore simple data augmentation, mixup, and adverserial attacks for creating perturbed distributions. We find that the margin distribution summary on mixed-up samples is most predictive of generalization.  

\section{Experiments}

We evaluate our methods on the 8 datasets provided as part of the PGDL Competition. The public datasets of this competition contain 150 deep CNN models trained on the CIFAR-10 and SVHN datasets. The CIFAR-10 models are VGG-like, while the SVHN models follow a Network-in Network type of architecture. These networks are trained with various hyperparameter choices - convolutional width, number of layers, batch size, dropout, depth, and weight decay. All models are trained till saturation, i.e. nearly perfect training accuracy, while their test performance varies.  



\renewcommand{\arraystretch}{1.3}

\begin{table}[h]
\centering
\begin{tabular}{@{}lclcl@{}}
\toprule
\textbf{Complexity Measure}       &  & \textbf{CIFAR-10} &  & \textbf{SVHN} \\ \hline
                                  &  &                   &  &               \\
Davies Bouldin Index              &  & 1.76 (25.22)      &  & 7.98 (22.19)  \\
Label-Wise Mixup                  &  & 0.00              &  & 22.75         \\
Margin Summary                    &  & 1.87              &  & 39.95         \\
Manifold Mixup                    &  & 2.24              &  & 12.11         \\
DBI * Label-Wise Mixup &  & 0.00              &  & 32.05         \\
Augment Margin Summary            &  & 5.73 (15.66)             &  & 44.60 (48.34)        \\
Mixup Margin Summary              &  & 1.11              &  & 47.33         \\
prod-of-spec-over-margin            &  & 1.54              &  & 5.13         \\
prod-of-fro-over-margin            &  & 2.84              &  & 4.92         \\
Sharpness                       &  & 1.77              &  & 15.85        
\\
Augment Performance                       &  & 1.54              &  & 6.61         \\
VC-Dimension                       &  & 0.00              &  & 0.00         \\
\bottomrule
\end{tabular}
\vspace{1 em}
\caption{Mutual Information Scores for various representation based complexity measures as well as some baseline measures on CIFAR-10 and SVHN. Specific details for the measures as well as design choices for the task-specialized scores in brackets are given in the Appendix.}
\label{tab: table 1}
\end{table}

\begin{table}
\centering
\begin{tabular}{lllll}
\toprule
\textbf{Complexity Measure}       & \textbf{CIFAR-10} & \textbf{SVHN} & \textbf{PGDL Task 4} & \textbf{PGDL Task 5} \\ \hline
                                  &                   &               &                      &                      \\
DBI * Label-Wise Mixup & 0.00              & 32.05         & 31.79                & 15.92                \\
Augment Margin Summary            & 5.73              & 44.60         & 47.22                & 22.82                \\
Mixup Margin Summary              & 1.11              & 47.33         & 43.22                & 34.57   \\
\bottomrule
\end{tabular}
\vspace{1 em}
\caption{Mutual Information scores of our final complexity measures on public and development tasks of PGDL}
\label{tab: table 2}
\end{table}

\begin{table}
\centering
\begin{tabular}{lllll}
\toprule
\textbf{Complexity Measure}       & \textbf{PGDL Task 6} & \textbf{PGDL Task 7} & \textbf{PGDL Task 8} & \textbf{PGDL Task 9} \\ \hline
                                  &                      &                      &                      &                      \\
DBI * Label-Wise Mixup & 43.99                & 12.59                & 9.24                 & 25.86                \\
Augment Margin Summary            & 8.67                 & 11.97                & 1.28                 & 15.25                \\
Mixup Margin Summary              & 11.46                 & 21.98                & 1.48                & 20.78  \\
\bottomrule
\end{tabular}
\vspace{1 em}
\caption{Mutual Information scores of our final complexity measures on final tasks of PGDL}
\label{tab: table 3}
\end{table}

Tables \ref{tab: table 1}, \ref{tab: table 2} and \ref{tab: table 3} show the performance of our methods on the different tasks of PGDL. We report scores of individual complexity measures as well as combinations which achieve highest scores. Trivially combining measures is difficult without access to the distribution of measures over the entire task; scale differences between non-normalized measures across models makes it difficult to create meaningful combinations. Our highest scoring complexity measure on the final task combines two measures - DB Index and Label-Wise Mixup, both of which have the advantage of being relative quantities and are hence scale invariant, allowing us to combine them by simply taking a product. We also report scores of some baseline measures taken from Jiang et al. (2019) \citep{jiang2019fantastic}. For some complexity measures, various design choices are possible and lead to high scores on individual tasks. In such cases we show the scores of generic/minimalist variants of each measure which we expect to run across tasks irrespective of type of data and model architectures, as well as the highest scores obtained by making intelligent design choices based on knowledge of the data and model architectures (shown in brackets in Table \ref{tab: table 1}), which is achievable in a real-world setting. We find that calculating our complexity measures over as little as 1\% of the training samples is enough to give reasonable estimates of generalization ability, underlining their computational efficiency. Details about design choices are detailed in the appendix. We show the scores of all complexity measures we tested on the public tasks; however, due to limited access to the private tasks we only show the scores of our final 3 solutions on these.

\section{Discussion}

A number of methods have been proposed before for understanding and predicting generalization performance in deep learning. However, many of these are in the form of upper bounds which are unsuitable to use in a practical setting, while others are unreliable when dealing with a wide range of hyperparameters and topological differences in models.

In this work, we have introduced complexity measures based on the quality of representations of deep neural networks that are especially useful in a post-hoc setting. We discuss our motivation and approach that led to a top score in the NeurIPS Competition on Predicting Generalization in Deep Learning. We demonstrate the effectiveness and ubiquity of our measures on a wide range of tasks part of the PGDL Competition. We show that our measures can lead to better mutual information scores than baseline measures such as spectral norm and sharpness on CIFAR-10 and SVHN datasets. Our main observations and conclusions are as follows:

\begin{itemize}
    \item Simple measures based on the quality of internal representation are predictive of generalization gap. Specifically, we demonstrate how quantifying three qualities of representations - consistency, robustness, and separability, can predict generalization across a wide range of hyperparameter and model architectures.
    \item Our hypothesis aligns with neuroscientific theories of how the human visual system creates invariant object representations in the face of controlled naturalistic variations and has the ability to untangle inseparable representation manifolds into easily separable ones.
    \item We measure consistency of internal representations using the Davies Bouldin Index. This indicates that clustered and compact representations of objects in internal layers aid in generalization.
    \item To measure robustness, we use a label-wise variation of Mixup. Mixup as a technique is uniquely qualified to probe the invariance of representations inside the model in a post-hoc setting since it is both data-agnostic and controlled, i.e. it does not introduce external elements which might take the model into unseen regions.
    \item We measure separability of representations using the Margin Distribution, which has been explored before. However, we find that the perturbed margin is more expressive of generalization than just the margin.
\end{itemize}

While it is not surprising that better representations lead to better generalization, there have been limited studies to explicitly understand the relationship between these. We provide a way to think of generalization from the perspective of the parallels of deep neural network representations with the human visual system. We show that interpreted this way, existing regularization techniques can be modified to be more suitable for predicting generalization across a wide range of tasks in a post-hoc setting. We hope our methods and analysis can provoke further research on understanding qualities of representation and other intermediate measures that are responsible for better generalization, as well as more critical evaluations of the parallels between deep neural networks and the human visual system.

\section{Acknowledgements}

We'd like to thank the organizers of the NeurIPS 2020 Competition on Predicting Generalization in Deep Learning for hosting this competition and for providing a platform for us to test our hypotheses. 

\bibliographystyle{plainnat}
\bibliography{neurips_edu_2020}

\section{Appendix}
\label{sec : appendix}

\subsection{Details for Table \ref{tab: table 1}}

Here, we detail the various design choices for complexity measures in Table \ref{tab: table 1}.

\begin{enumerate}
    \item For the generic version of the Davies Bouldin Index, we use max-pooling with a filter size of 4 and stride of 1 for spatial aggregation/dimensionality reduction for better clustering. We compute the DB Index on the first layer, since this is most suitable when model architectures are unknown. For the task-specialized version for CIFAR-10 and SVHN, we compute DB Index on the third-from-last layer. This may be a max-pool layer or a convolutional layer for different models; we find that comparing across these layer types still gives reasonable mutual information scores.
    \item For Label-Wise Mixup, we hold $\lambda$ constant at 0.5. 
    \item For the Margin Summary, we use the mean of the quantile signature in Jiang et al. (2018) \citep{jiang2018predicting}. We also explore simply the mean of the margin distribution and find that this gives similar results.
    \item Manifold Mixup scores are reported on the first convolutional layer since we are comparing models with varying topologies; we do not conduct a study to explore the suitability of different intermediate layers for this measure.
    \item For the Augment Margin Summary, we use the following augmentations: Random Hue with a maximum value of 0.5, random saturation with a saturation value sampled from [0.6, 1.2], random brightness with a maximum value of 0.5, random contrast sampled from [0.7, 1.0], random clipping/zoom [1 to 15 \%], and left-right flips. For the generic version, we only use contrast, left-right flips, and zoom. Random seeds are fixed so that all models are tested on the same vicinal distribution.
    \item Both norm-based measures, prod-of-spec-over margin and prod-of-fro-over-margin, are computed as in Jiang et al. (2019) \citep{jiang2019fantastic}.
    \item For Augment Performance, we simply check the categorical cross-entropy of the model on augmented samples. Augmentations are same as those for the Augmented Margin.
\end{enumerate}


\small





\end{document}